\documentclass{llncs}

\usepackage{gensymb}
\usepackage{mathtools}
\usepackage{adjustbox}
\usepackage{multirow}
\usepackage{booktabs,subcaption,amsfonts,dcolumn}
\usepackage{caption}
\usepackage{subcaption}
\usepackage{xcolor}
\usepackage{url}
\usepackage{epstopdf}

\usepackage{hyperref}

\makeatletter
\newcommand{\printfnsymbol}[1]{%
  \textsuperscript{\@fnsymbol{#1}}%
}
\makeatother

\begin{document}
\title{People Tracking in Panoramic Video\\for Guiding Robots}
\author{Alberto Bacchin\thanks{The authors equally contribute to this paper} \and Filippo Berno\printfnsymbol{1} \and Emanuele Menegatti \and Alberto Pretto
\institute{All authors are with the Intelligent Autonomous Systems Laboratory (IAS-Lab),
Dept. of Information Engineering, University of Padova,
Via Gradenigo 6/B, 35131 Padova, Italy\\
\email{\{bacchinalb,emg,alberto.pretto\}@dei.unipd.it}}
}

\maketitle
\begin{abstract}
A guiding robot aims to effectively bring people to and from specific places within environments that are possibly unknown to them. During this operation the robot should be able to detect and track the accompanied person, trying never to lose sight of her/him. A solution to minimize this event is to use an omnidirectional camera: its 360$\degree$ Field of View (FoV) guarantees that any framed object cannot leave the FoV if not occluded or very far from the sensor. However, the acquired panoramic videos introduce new challenges in perception tasks such as people detection and tracking, including the large size of the images to be processed, the distortion effects introduced by the cylindrical projection and the periodic nature of panoramic images. In this paper, we propose a set of targeted methods that allow to effectively adapt to panoramic videos a standard people detection and tracking pipeline originally designed for perspective cameras. Our methods have been implemented and tested inside a deep learning-based people detection and tracking framework with a commercial 360$\degree$ camera. Experiments performed on datasets specifically acquired for guiding robot applications and on a real service robot show the effectiveness of the proposed approach over other state-of-the-art systems.  We release with this paper the acquired and annotated datasets and the open-source implementation of our method.
\end{abstract}

\section{Introduction}
\label{intro}

\begin{figure}[t!]
    \centering
    \begin{subfigure}[b]{0.285\textwidth}
    \centering
    \includegraphics[width=0.95\textwidth]{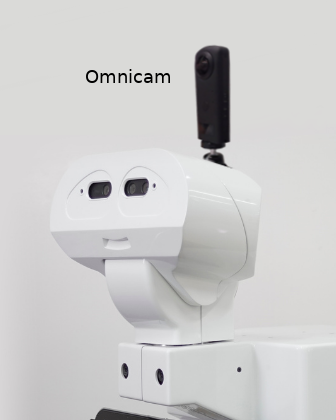}
    \label{fig:omni_bot}
    \end{subfigure}
    \hfill
    \begin{subfigure}[b]{0.705\textwidth}
        \centering
        \includegraphics[width=0.95\textwidth]{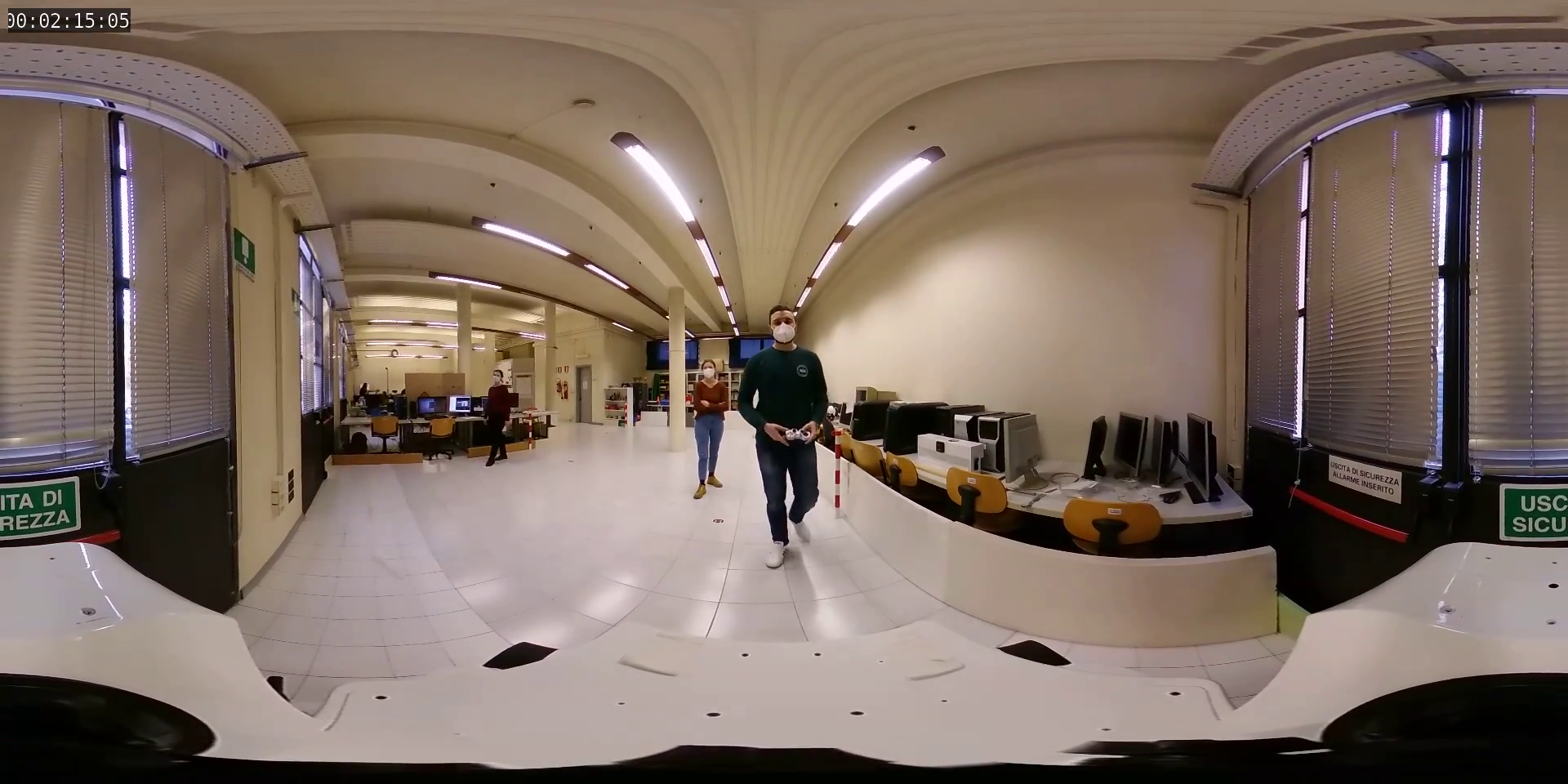}
        \label{fig:equirectangular}
    \end{subfigure}
    
\caption{(Left): The dual–fisheye omnidirectional camera mounted on the robot used in the experiments; (Right): an example of equirectangular panoramic image acquired during the experiments. This image is obtained by applying a cylindrical projection to the two images acquired by the fisheye cameras and by a subsequent image stitching phase.}
\label{fig:omnicam}
\end{figure}

Guiding robots were one of the first examples of the successful application of autonomous mobile robots outside the domain of academic research \cite{BURGARD19993}. In large public spaces like museums, hospitals, hotels, and many others, a guiding robot can help people to reach the desired destination by accompanying them. To achieve this behavior, it is important to be able to detect and track people in the robot space, even if they are temporarily occluded. 
The robot's on-board cameras usually have a reduced Field of View (FoV) which limits the tracking capabilities since the target person can easily get out of the FoV, for example when the robot is turning.
Omnidirectional cameras can solve this problem thanks to the 360$\degree$  FoV which always frames the whole surrounding environment. Nowadays several low-cost, lightweight, and high-quality omnidirectional cameras (also called 360$\degree$ cameras) are available in the market (e.g., left part of Fig. \ref{fig:omnicam}), enabling plenty of new applications and challenges for mobile robots.
In this work, we target the use of an omnidirectional camera for people detection and tracking, applied to the specific domain of guiding robots. We build upon the monocular people following framework proposed by Koide \emph{et al.} \cite{KOIDE2020103348}, we call here \textit{MonoPTrack}, adapting it to equirectangular panoramic videos (e.g., right part of Fig. \ref{fig:omnicam}).
\textit{MonoPTrack} is based on \textit{OpenPose} \cite{openpose}, a popular deep neural network for human skeleton detection. The panoramic images provided by an omnidirectional camera could be directly supplied as input to this system.
Unfortunately, due to the high computing resources required, \textit{OpenPose} is unable to perform inference in real-time on full resolution panoramic images on standard GPUs (Graphics Processing Units). On the other hand, performing inference on a downscaled image can lead to false-negative detections or wrong person re-identification. Moreover, the 360$\degree$ FoV comes at the cost of severe image distortions. In our case, the dimensions of a person in the image decrease very rapidly with the distance (e.g., right part of Fig. \ref{fig:omnicam}) so that in a downscaled image \textit{OpenPose}
cannot detect anyone more than 3-4 meters away from the camera. Furthermore, when mapping the entire surrounding scene to an equirectangular image, a \textit{horizontal wrap-around connection} is introduced.
The wrap-around structure of panoramic images is made clear when a target subject (e.g., a person) exits from one side of the image and reappears at the same time on the opposite side. The phenomenon easily leads traditional tracking strategies to failure.\\\\
In this paper we propose specific solutions to address the aforementioned problems.
In particular, we faced the challenges deriving from the high resolution of panoramic images and from the distortion introduced by cylindrical projection by introducing two simple but effective strategies, called \textit{ROI} and \textit{TILES}, that boost the detection and re-identification performance of existing people pose estimation systems. We also introduced a tracking strategy that allows to explicitly address the wrap-around structure of panoramic images. 
We evaluated the proposed techniques on a real world dataset, showing the benefits in terms of detection robustness, efficiency, and tracking accuracy.\\
The main contributions of this paper are: 
\begin{itemize}
    \item Two new strategies for addressing the challenges associated with detecting people in panoramic video such as high resolutions and projective distortion, enabling the use of omnidirectional cameras on real-time robotic applications;
    \item An effective solution for tracking people in 3D space starting from the equirectangular representation provided by an omnidirectional camera;
    \item A comparison of the proposed method against a current state-of-the-art method;
    \item An open-source implementation of our system and the acquired datasets are made publicly available with this paper:
    \begin{center}
    \url{https://github.com/bach05/PanoramicVideoPeopleTracking.git}    
    \end{center}
    
\end{itemize}

The rest of the paper is organized as follows. Sec. \ref{review} contains a review of related works. In Sec. \ref{methods}, we introduce the proposed \textit{ROI} and \textit{TILES} algorithms and the modification to a conventional tracking system to cope with panoramic videos. Sec. \ref{results} describes the evaluation procedure, the experiments performed, and a discussion of the results. Finally, conclusions and future research directions are provided in Sec. \ref{conclusions}.

\section{Related Work}
\label{review}

\subsection{People Detection}

Traditional approaches for detecting people rely on searching for a candidate region using a sliding window approach.
For each region, a set of features are extracted and used to classify it. In \cite{Aguilar}, authors used HAAR-LBP and HOG cascade classifier to detect people in surveillance videos. Since scanning the whole image multiple times with a sliding window approach is not so efficient, Geronimo \emph{et al.} \cite{pedestrian} proposed to use a calibrated camera to determine which regions are placed on the ground and so suitable for containing a person. 

However, nowadays deep learning-based methods have overcome the traditional approaches \cite{JI2020471}. Zhao \emph{et al.} \cite{zhao} proposed a Faster R-CNN architecture optimized for people detections. The Faster R-CNN \cite{fastercnn} architecture is inspired by the traditional object detection pipeline. In fact, it uses a Region Proposal Network to identify the regions of interest in the image and a classification network to assign the correct label to each proposed region. Other approaches are based on YOLO \cite{yolo}, another very popular object detector based on deep learning. In \cite{yolo4people}, authors used a YOLO person detector to locate people in the scene, relying on the centroid of the bounding boxes. Once the person is detected, through a calibrated camera, it is also possible to reconstruct her/his 3D position \cite{sharma2018pixels,Niu}.

Anyway, for human detection, bounding boxes do not always provide enough information. For these reason, human pose or skeleton estimation have been proposed \cite{openpose,kreiss2021openpifpaf}. In this case, the aim is to predict the position of N human body joints. This richer representation allows more precise localization in 3D space \cite{KOIDE2020103348,bertoni2019monoloco} without need of additional hardware such as RGB-D cameras \cite{KOLLMITZ201929,munaro} and laser range-finders \cite{Zheng}. 

\subsection{People Tracking}
\label{people_tracking}

The tracking task consists in associating consecutive detections of the same object. Optical flow can be used to estimate the motion of the objects for tracking purposes \cite{choi}. However, being a pixel-level computation, optical flow is not enough reliable and it works only in the image space. Instead, it is common to use an efficient state estimation algorithm, such as a Kalman Filter, to track detections. Ardiyanto \emph{et al.} \cite{ARDIYANTO2014904} exploited an Unscented Kalman Filter (UKF) to track people pose in the 3D world space. 
Using only the pose may not be robust enough, especially in crowded environments where trajectories are close and often intercept.
Adding visual features for data association have been demonstrated to be more robust \cite{LIU202018}. Also, Deep Learning has been explored for tracking \cite{deeptracking}. 

Despite over the years there was a huge interest in people detection and tracking, only a few works focused on the possibility of exploiting panoramic videos. In \cite{Thaler}, authors propose a people tracking system on panoramic videos obtained from a system of multiple cameras. They focused on real-time processing, relying on a HOG detector and an optical flow detector tracker. A more advanced approach, based on Siamese Networks, is proposed by Tai \emph{et al.} \cite{tai}. Li \emph{et al.} \cite{li} instead used YOLOv3 and ResNet50 to detect and track people in panoramic videos. However, all of them are limited to the image space and so the applicability in robotics applications remains unexplored. Recent studies used a couple of calibrated omnidirectional cameras to achieve accurate 3D people tracking in panoramic videos \cite{Shere}. The proposal of Yang \emph{et al.} \cite{yang}, called \textit{Multi-person Panoramic Localization and Tracking (MPLT)}, instead relies on four calibrated perspective cameras, positioned such that they cover a field of view of 360$\degree$ to simulate an omnidirectional camera. The OpenPifPaf framework \cite{kreiss2021openpifpaf} is used for skeleton detection and a pinhole camera model is used to map detections in the 3D space, under the assumption that the height of a person is almost constant. The tracking system models each couple of tracks with two costs, one associated with the appearance of the person and one associated with the distance of the trajectories estimated with a Kalman Filter. The tracking is performed by finding the associations that minimize the costs. However, compared to our approach, multi-camera configurations require more complex calibration processes and higher computational costs.

\section{Methodology}
\label{methods}
\subsection{Baseline}
\label{baseline}
 
\begin{figure}[t!]
    \centering
    \includegraphics[width=1.0\textwidth]{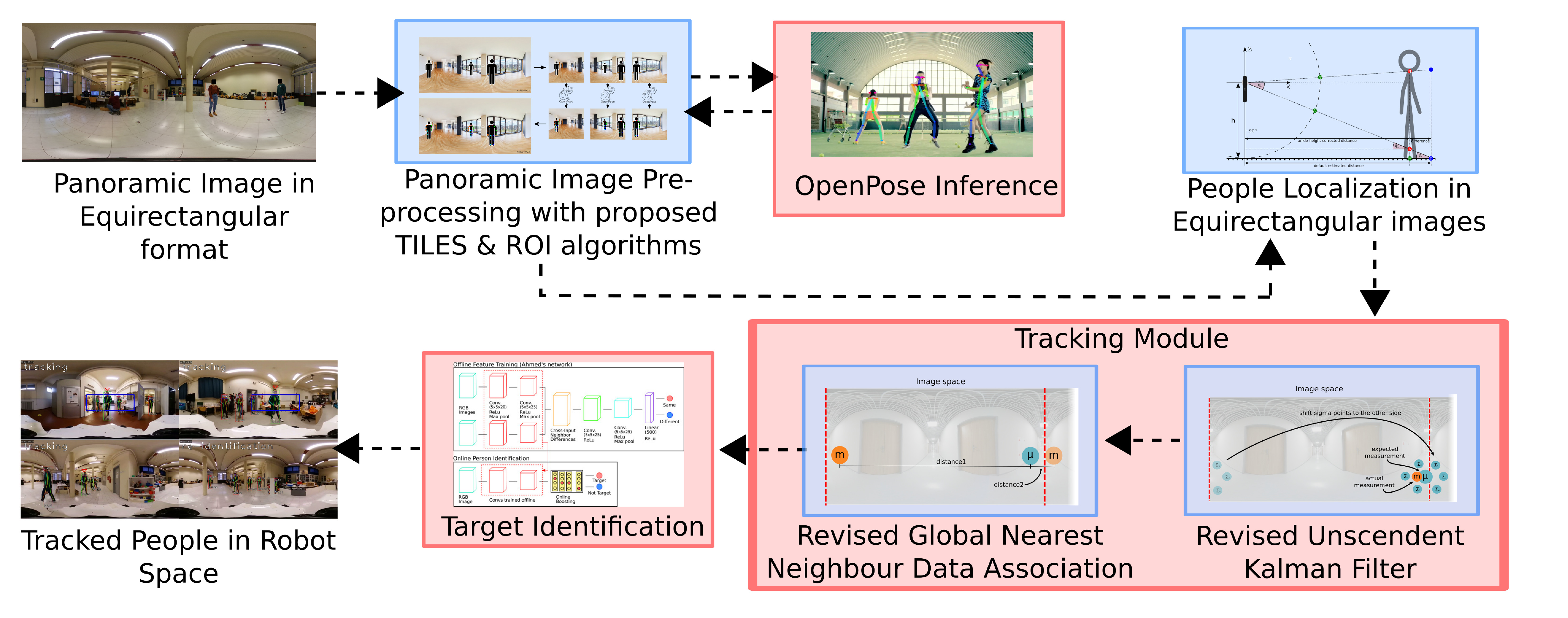}
    \caption{Block diagram of the proposed pipeline. Blue blocks highlight the aspects we contribute on, red blocks instead are taken from previous methods.}
    \label{fig:scheme}
\end{figure}

Our system is based on the \textit{MonoPTrack} framework \cite{KOIDE2020103348}, with new specific strategies aimed at improving the robustness of the person detection front-end and the target tracking sub-system. Thanks to parallel pipelines and small-sized processed images, the proposed algorithms also made it possible to improve runtime performance even on low-end computers, such as those that typically equip service robots. The front-end of \textit{MonoPTrack} is represented by the OpenPose \cite{openpose} skeleton detection. The neck and ankles joints are used to estimate the position of a person in the 3D space, given a calibrated camera. An UKF is used in the tracking process. UKF \cite{ukf} is a sampling based-method designed to overcome the limitation of a traditional Kalman Filter that works only for linear models. The idea is to select a set of representative points from the starting distribution, called \textit{Sigma Points}, apply the non-linear transformation to each point, and finally estimate a Gaussian distribution from the sampled points. Tracking is completed with a Global Nearest Neighbour (GNN) data association. The last module of the original pipeline is the re-identification module which learns appearance features of the target person. Our work focused on three points of the original pipeline. (i) Before skeleton detection, we introduced the \textit{ROI} and \textit{TILES} pre-processing algorithms to ensure better runtime performance while boosting the detection rate also in case of people far from the camera, hence projected in a small region of the panoramic image. (ii) We modified the camera model to deal with equirectangular panoramic images in order to correctly map from image space to robot space; here we also introduced ankle height correction, to improve the accuracy of person localization. (iii) Finally, we modified the tracker to cope with the \textit{wrap-around} connections. A scheme of the proposed pipeline can be found in Fig. \ref{fig:scheme}. 

\subsection{The TILES approach}
\label{tiles}

    
    
\begin{figure}[t!]

    \centering
    \begin{subfigure}[b]{0.9\textwidth}
    \centering
    \includegraphics[width=0.85\textwidth]{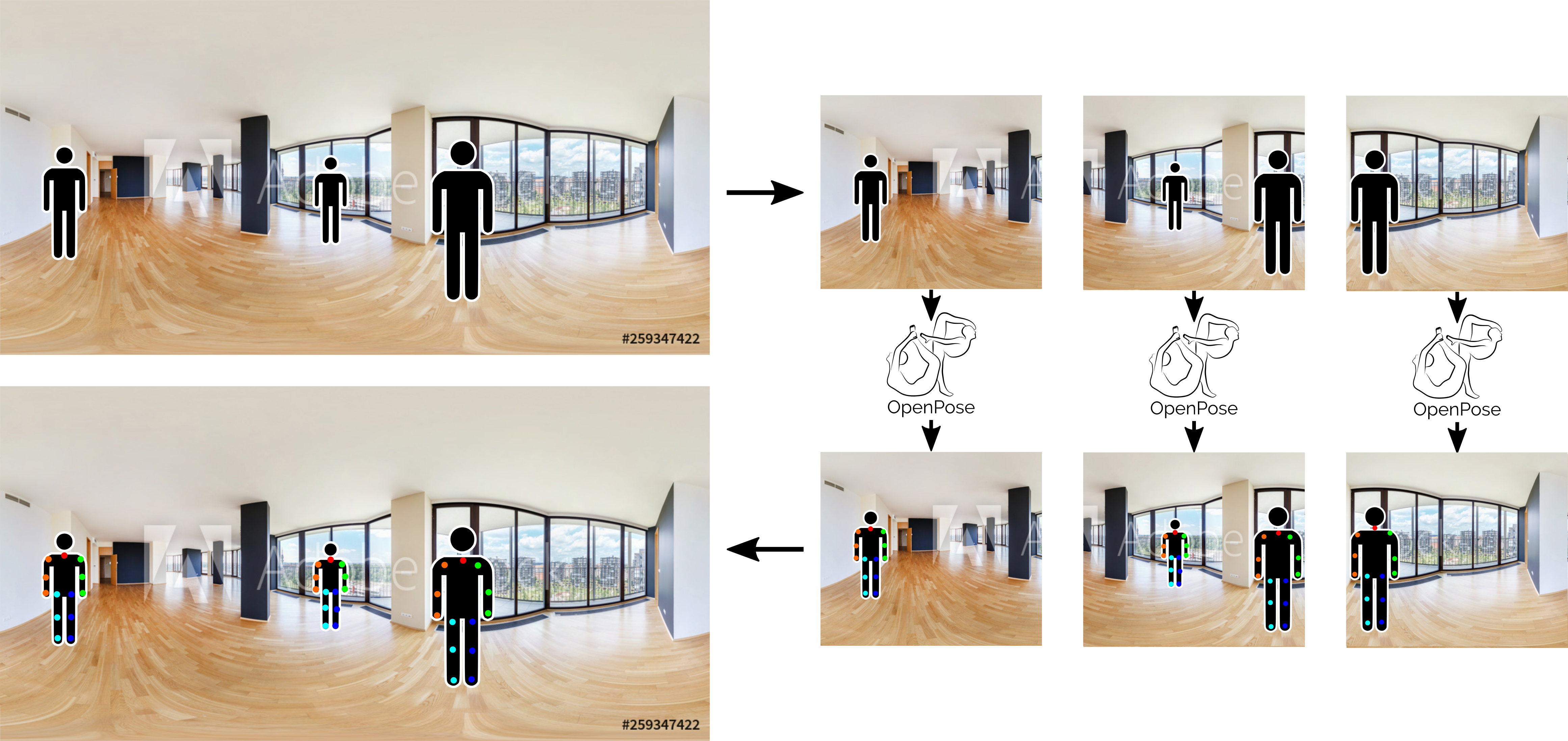}
    \caption{Graphical representation of the \textit{TILES} approach. Each tile is processed by a dedicate instance of \textit{OpenPose} and then all the detections are fused. }
    \label{fig:tiles_proc}
    \end{subfigure}
    \hfill
    
    \begin{subfigure}[b]{0.9\textwidth}
        \centering
        \includegraphics[width=0.85\textwidth]{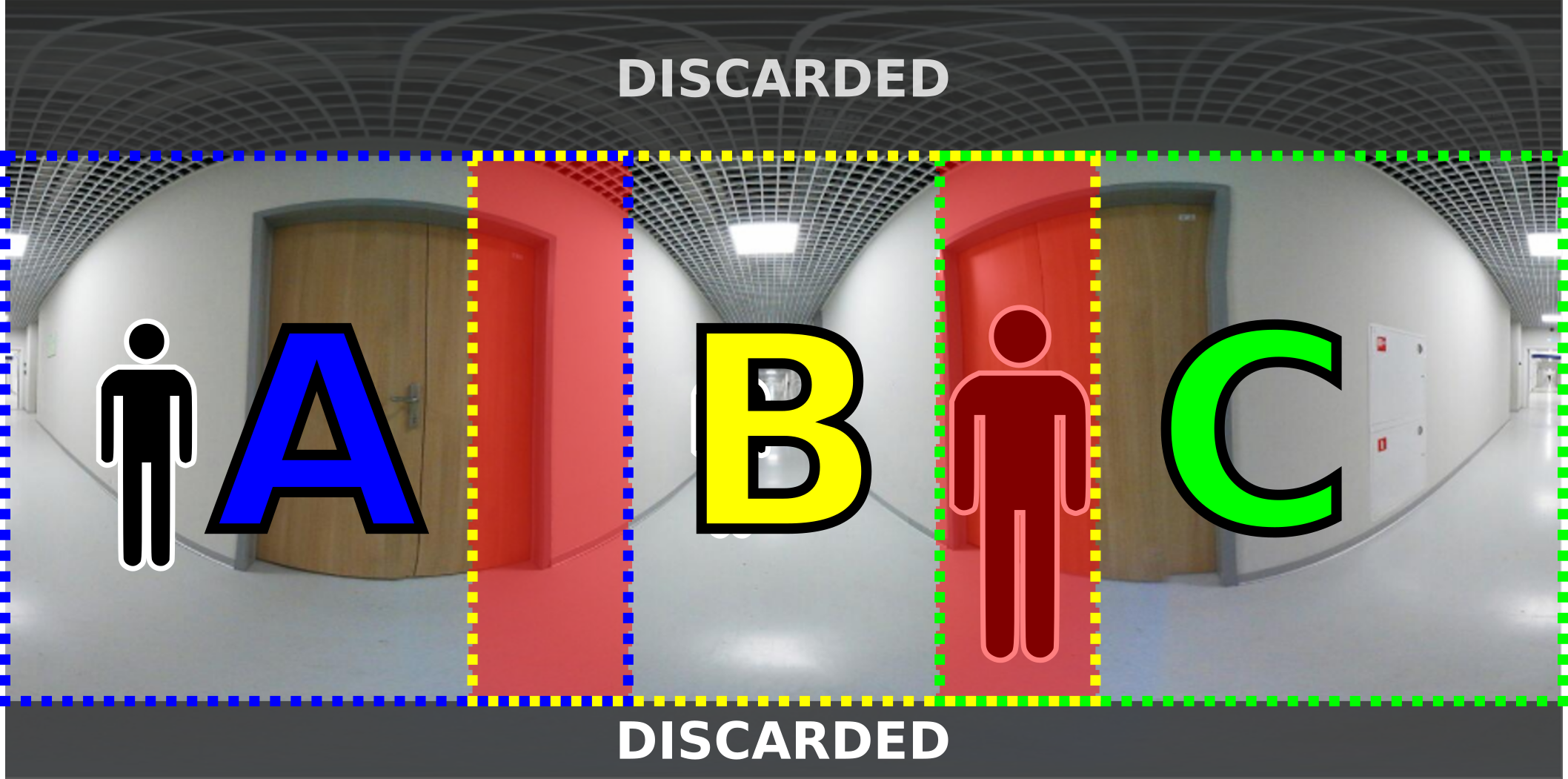}
        \caption{The subdivision of the image into tiles; the upper and lower zones are discarded because no useful information is contained there. The three tiles are marked with A, B and C, while overlapping zones are in red.}
        \label{fig:tiles_division}
    \end{subfigure}
    
\caption{The working principle of the \textit{TILES} method.}
\label{fig:tiles}
\end{figure}

The \textit{TILES} algorithm focuses on the problem of efficiently inferring skeleton detection from high-resolution images. The aim is to reduce the resolution while parallelizing the inference. We divide the image into three equal parts. Each portion, namely tile, will be analyzed by a dedicated instance of \textit{OpenPose} and results will be merged. The division can split a person over two consecutive tiles, preventing the detection. To solve this issue, tiles are designed to be slightly overlapping, as shown in Fig. \ref{fig:tiles}. The overlapping area is 150-pixel wide, which is the width of a person 1 meter away from the robot. We take 1 meter as worst case since is rare that a person stays closer to a moving robot. 

A person in the overlapping area can be detected twice. To avoid double detection, before delivering the set of detections, we compute a bounding box containing shoulder, neck, and hip joints around each person. For each pair of adjacent tiles and each pair of bounding boxes $(B_1, B_2)$, we compute the following score:
\begin{equation}
S(B_1, B_2) = \frac{| B_1 \cap B_2 |}{|min(B_1, B_2)|} \label{eq:merge_score}
\end{equation}
If $S(B_1, B_2) \ge \sigma_1 $, we merge $B_1$ and $B_2$ in a single detection. People at different distances from the camera and with different physical appearances (i.e. different joint positions) will be bounded with different sized boxes. This strategy allows fusing double detection with a low overhead since it does not need additional information like extracting visual features from people. 

\subsection{The ROI approach}

\begin{figure}[t]
    \centering
    \includegraphics[width=0.85\textwidth]{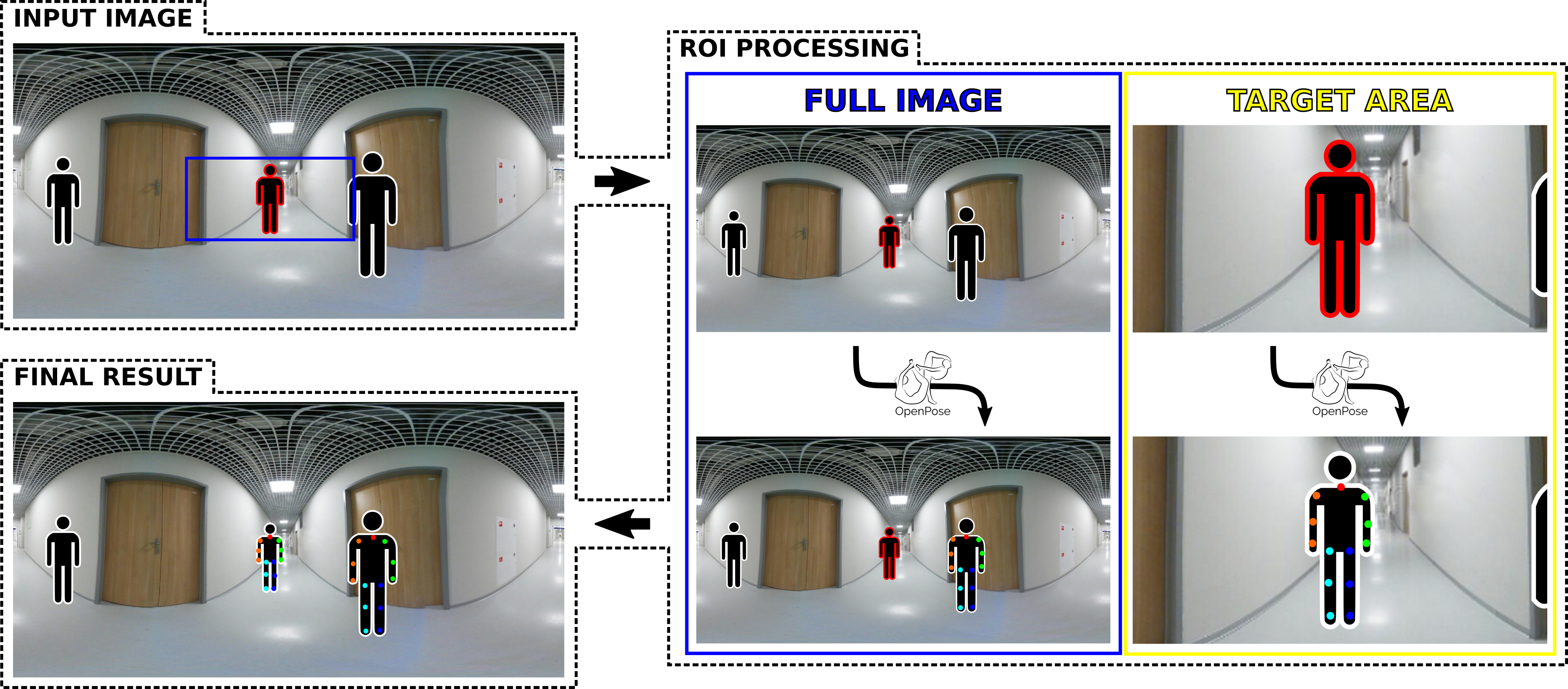}
    \caption{The image shows the working principle of the \textit{ROI} method. One instance of \textit{OpenPose} processes a low-resolution version of the full image. Another processes a portion of the full image, containing the target person, outlined in red.}
    \label{fig:roi}
\end{figure}

The \textit{TILES} algorithm, thanks to the parallelization, provides a good processing speed while improving detection accuracy of further people. The \textit{ROI} (i.e., \emph{Region Of Interest}) approach explores the possibility of further improvements, under the specific case of a guidance robot that usually focuses on a specific target person. The first frame is re-scaled and processed only with a single open \textit{OpenPose} inference to obtain the target. Identifying a meaningful target person is beyond the scope of this work, so we promote as target the first detected person. Then, \textit{ROI} performs two \textit{OpenPose} inferences for each frame, see Fig. \ref{fig:roi}. The first on the whole panoramic image, but re-scaled to lower resolution. The second is performed on a specif region of interest, from the full-resolution image, which contains the target person indicated by the re-identification module on the previous frame. We use the same method illustrated in Sec. \ref{tiles} to merge double detections. Focusing on a specific region, the network can detect a person very far from the camera, at the cost of worse performances in multi-person detection because of the re-scaling. 

\subsection{People Tracking in Panoramic Videos}
\label{tracking}

\begin{figure}[h]
    \centering
    \includegraphics[width=0.8\textwidth]{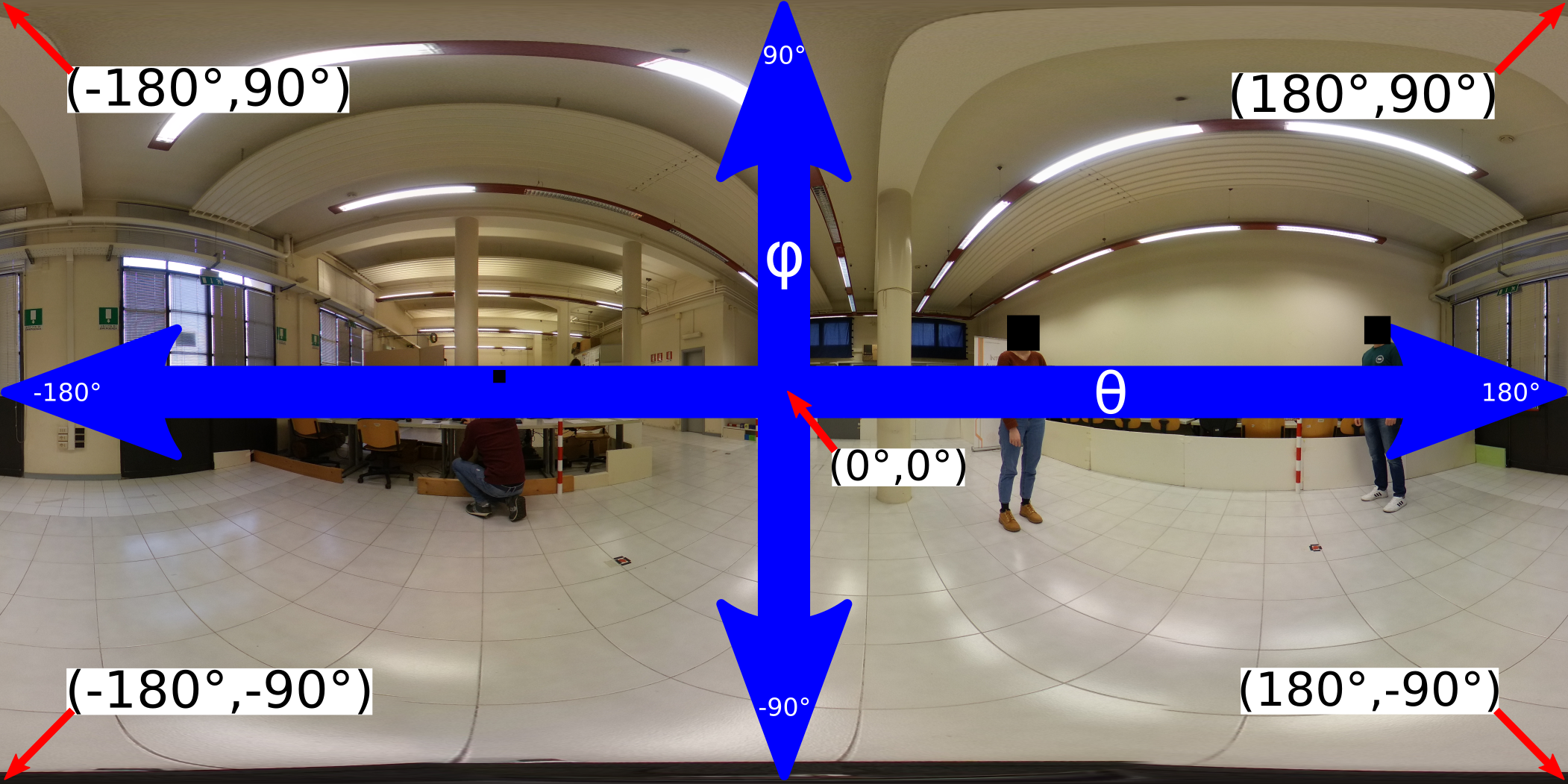}
    \caption{In this pictures, we can see how the equi-rectangular format is mapped into polar coordinates. The column determines the angle $\theta$, the row determines the angle $\varphi$.}
    \label{fig:polar_coordinates}
\end{figure}

Once skeleton detection is performed, we use the information of neck and ankles joints to estimate the location of each perceived person. This usually involves a calibrated camera, but we instead exploited the peculiar characteristics of the equirectangular image, assuming that the camera is positioned orthogonal to the ground at a fixed height. The equirectangular format is representing a sphere on a flat surface, so it is straightforward to use polar coordinates $(\theta, \varphi, \rho)$, as shown in Fig.\ref{fig:polar_coordinates}.  

Given the middle point between the two ankles joints ($\alpha$) in the 3D space, its polar coordinates can be computed from its image coordinates $(x_\alpha, y_\alpha)$: 

\begin{equation}
\begin{array}{l}
\theta_\alpha = 180 - \frac{FoV_H}{I_W}x_\alpha \;\;\;\;
\varphi_\alpha = 90 - \frac{FoV_V}{I_H}y_\alpha \;\;\;\;
\rho_\alpha = \frac{h-k}{tan(\varphi_\alpha \cdot \frac{\pi}{180})} \label{eq:dist}
\end{array}
\end{equation}

where $FoV_H$ and $FoV_V$ are horizontal and vertical fields of view in degrees, $I_W$ x $I_H$ is the panoramic image resolution in pixels, $h$ is the height of the camera from the ground and $k$ is the average ankle height from the ground for a human, both in meters. Thanks to the neck joint (N), it is possible to estimate the person height:

\begin{equation}
\begin{array}{l}
h_N = h + \rho_N \cdot tan(\varphi_N \cdot \frac{\pi}{180})
\end{array}
\end{equation}

The position of a person P in Euclidean coordinates is:

\begin{equation}
\begin{pmatrix}
X_P\\ 
Y_P\\ 
Z_P
\end{pmatrix} = \begin{pmatrix}
\rho_\alpha \cdot cos(\theta_\alpha)\\ 
\rho_\alpha \cdot sin(\theta_\alpha)\\ 
h_N
\end{pmatrix}
\end{equation}

\begin{figure}[h]
     \centering
     \begin{subfigure}[b]{0.33\textwidth}
         \centering
         \includegraphics[width=\textwidth]{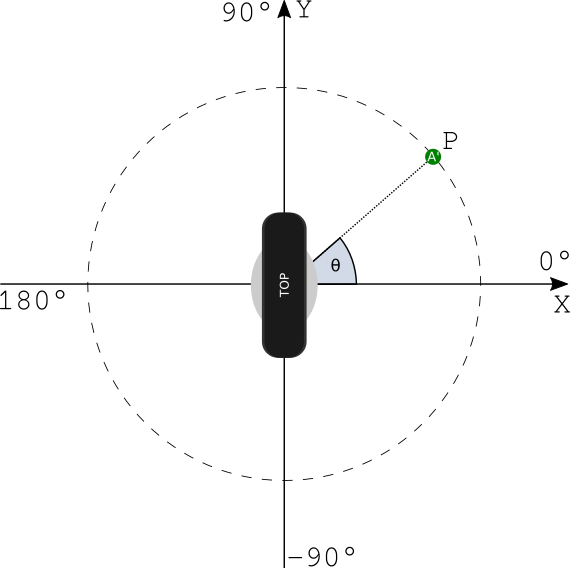}
         \caption{The system from above.}
         \label{fig:loc2}
     \end{subfigure}
     \hfill
     \begin{subfigure}[b]{0.57\textwidth}
         \centering
         \includegraphics[width=\textwidth]{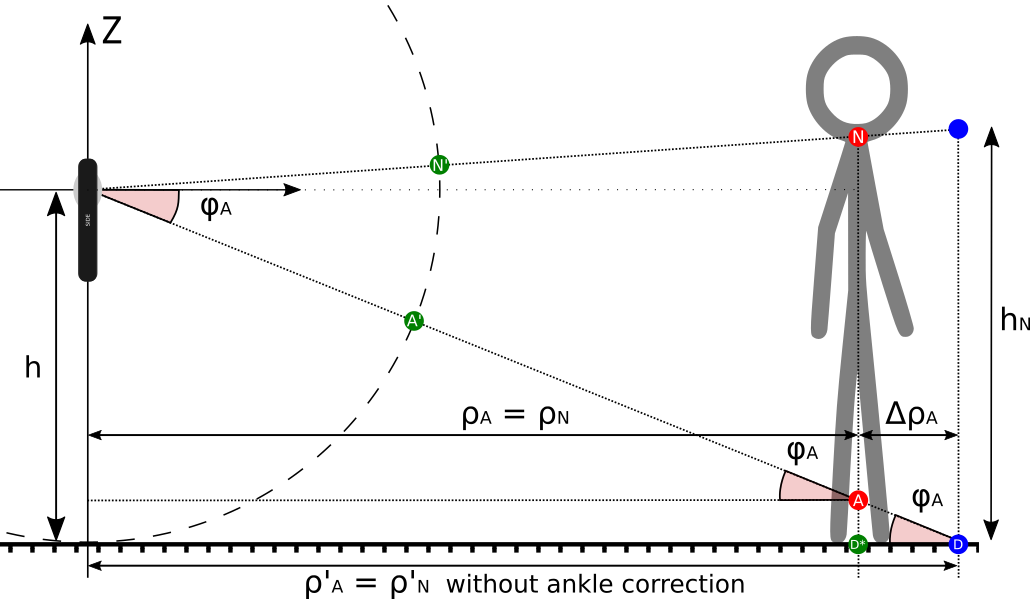}
         \caption{The system from the side.}
         \label{fig:loc1}
     \end{subfigure}
     \caption{The geometry behind the mapping from image space to 3D space. The dashed circumference represents the image plane of the omni-directional camera.  $\alpha'$ and $N'$ are the points in the image that corresponds to the ankle and neck joints. $\alpha$ and N are the corresponding points in the robot space. $h$ is the height of the camera above the ground. $h_N$ is the neck height which approximates person's height. On the right, it is also shown how the ankle height correction improves the final estimation of the person distance from the camera}
     \label{fig:loc}
\end{figure}

The neck joint and the height estimation in the state allow updating the filter even if the ankles are occluded for more stable tracking. Fig. \ref{fig:loc} shows visually the aforementioned geometric transformations. 

\begin{figure}[h]
    \centering
    \begin{subfigure}[b]{0.75\textwidth}
        \centering
       \includegraphics[width=\textwidth]{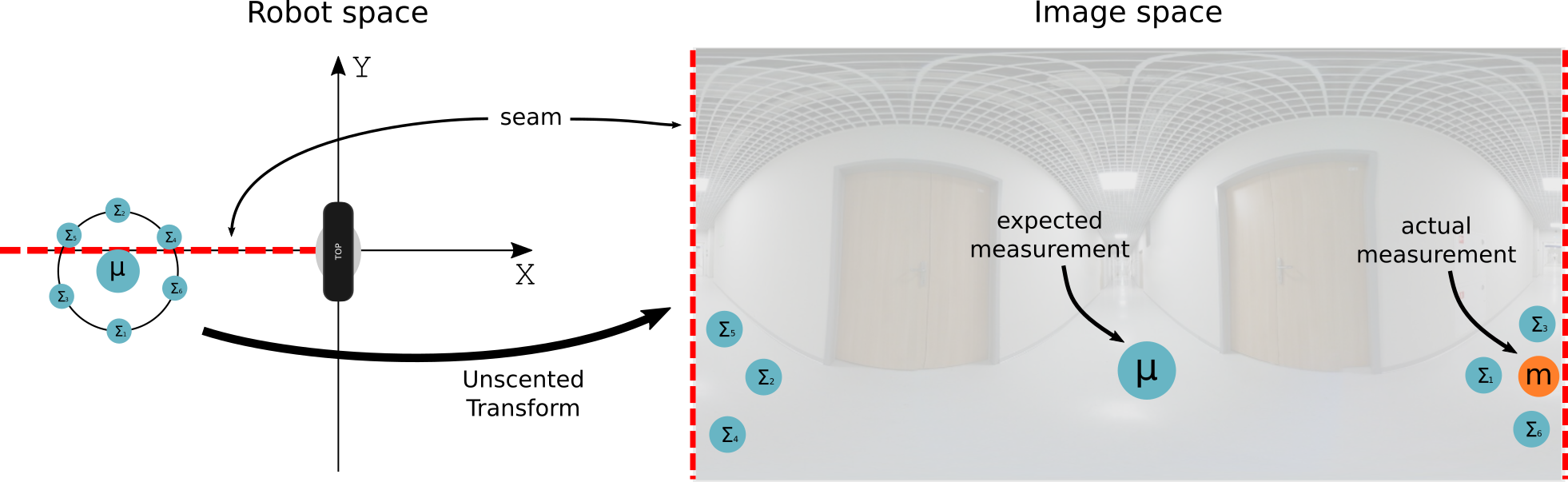}
        \caption{Wrong mean estimation due to wrap-around. }
        \label{fig:ut}
    \end{subfigure}
    
    \begin{subfigure}[b]{0.65\textwidth}
        \centering        \includegraphics[width=\textwidth]{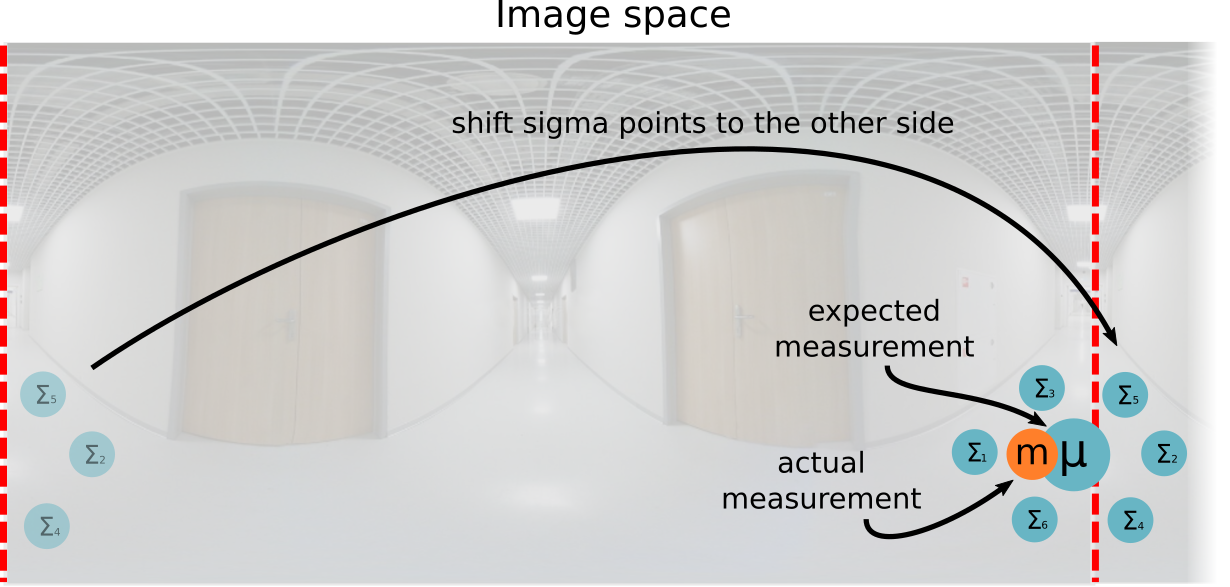}
        \caption{Corrected estimation.}
        \label{fig:ut_shift}
    \end{subfigure}
\caption{The figures shows the correction applied to the UKF. $\Sigma_i$ are the sigma points, $\mu$ is the mean of the sigma points. (a) Disruptive impact of Sigma points mapped on the other side of the image. (b) Correction applied to obtain the correct mean estimation.}
\label{fig:ukf_correction}
\end{figure}

The \textit{wrap-around} problem, introduced in Sec. \ref{intro}, forces to modify the UKF and the data association strategy in the tracking module. As explained in Sec. \ref{baseline}, UKF uses the \textit{Sigma points} to deal with non-linear functions. When a person is closed to the boundaries of the equirectangular image, it may happen that \textit{Sigma points} from the starting distribution in the image would be mapped on two different sides. This fact has disruptive effects on the estimation of the predicted distribution, see Fig. \ref{fig:ut}. To solve, we shift the \textit{Sigma points} fallen on the opposite side of the image by increasing their x-coordinate by an amount equal to the width of the image. This results in a correct average calculation, as shown in Fig. \ref{fig:ut_shift}. If the resulting average is outside the image space, it is sufficient to subtract the image width from the x coordinate.

Since the data association for the tracking takes place in the image space, the same problem related to the \textit{wrap-around} could arise. To tackle the issue, we reformulated the equation to compute the distance of two points, $P_1$ and $P_2$, in the image space: 

\begin{equation}
d(P_1,P_2) = \sqrt{(min(x_{P_1}-x_{P_2}, image_W - (x_{P_1}-x_{P_2})))^2+(y_{P_1}-y_{P_2})^2}
\end{equation}

where $I_W$ is the width of the image in pixels, $(x_{P_1}, y_{P_1})$ and $(x_{P_2}, y_{P_2})$ are the Cartesian coordinates of the points in the image space.

\section{Experimental Evaluation}
\label{results}

\subsection{Experimental Setup}

We collected a dataset of 14 videos of variable lengths, between 6 s and 134 s. Each frame can contain up to 5 people. The dataset was acquired with a Ricoh Theta Z1\footnote{\url{https://theta360.com/it/about/theta/z1.html}}, a dual-fisheye omnidirectional camera that allows a 360$\degree$ horizontal, 180$\degree$ vertical FoV and real-time streaming at 30 fps via USB connection. The rendered image is an equirectangular projection, generated by the camera itself, with a resolution of 1920x960 pixels. The recordings include both indoor and outdoor environments, with one or more people moving around the camera. In some videos the camera is static, in some others the camera was mounted on a mobile robot to simulate a guiding task. The distance from the ground and the camera sensor was 1.2 m.

The dataset images have been manually annotated with the neck and ankles joints of the target person, since we aim a guiding application where the robot focuses on a single person. The videos were not densely annotated, namely, we provided annotations every 10 or 20 frames which are enough for the evaluation of a tracking system. To avoid biases due to inaccurate assumptions, for example if the camera is not perfectly perpendicular, the ground truth of the target 3D position is computed using the model presented in Sec. \ref{tracking} from the annotation of neck and ankles joints on the images. We ended up with a total of 3200 annotated frames. A subset of the videos containing 1108 annotated frames was used for parameter tuning, the rest for the test.

We set a threshold $\theta_1 = 0.9$ for Eq. \ref{eq:merge_score} after experimental tuning. For \textit{ROI} algorithm, we set the size of the region of interest to 576x192 pixel, while the full image was re-scaled to 640x320 pixels. We used default parameters for the UKF and the tracker. 

The evaluation was performed offline on the pre-recorded test set with a PC equipped with an NVIDIA GTX 1650 GPU, 16 GB of RAM, and an Intel i7-4790 CPU. We also compared the performance of our proposal with the \textit{Multi-person Panoramic Localization and Tracking (MPLT)} \cite{yang} introduced in Sec. \ref{people_tracking}. Since \textit{MPLT} works with four calibrated cameras to provide a panoramic view, not directly with omnidirectional cameras, we split the equirectangular image into four portions of size 480 x 960 pixels. Each portion is meant to cover the area in front, behind, at the left, and the right of the camera, without overlapping. This approximates quite well the original configuration. Moreover, the code provided by the authors tries to load in memory the whole video before processing it, with a consequent saturation of the RAM with our dataset. We have reorganized the processing pipeline to elaborate the video frame by frame, as in real-time systems like our proposal. We have used the proposed camera model, adjusting the parameters for our camera. 

\subsection{Evaluation Metrics}

To design the evaluation metrics, we started from ETISEO metrics \cite{etiseo}. Given the practical application of a guiding robot, we propose three metrics that mainly focus on the performance in target tracking. 

The Normalized Detection Time, called $M1$, measures the percentage of time during which the target is detected by the system and is given by the following equation:

\begin{equation}
M1(v) = \frac{1}{N_v}\sum_{i=1}^{N_v} \frac{T_i}{F_i} \label{eq:M1}
\end{equation}

where $N_v$ is the number of people annotated in the video $v$ ($N_v=1$ in our case), $T_i$ is the number of frames in which person $i$ is tracked, $F_i$ is the number of frames for which person $i$ is annotated. $M1$ takes values between 0 and 1, high values are better.

$M2$, the Tracking Turnover, evaluates the persistence of the ID assigned to the target. This allows testing the continuity of the target tracking, since every time the tracking is lost and then restarted a new ID is assigned. $M2$ is defined by the following formula:

\begin{equation}
M2(v) = \frac{1}{N_v}\sum_{i=1}^{N_v} \frac{1}{Frag_i} \label{eq:M2}
\end{equation}

where $N_v$ is the number of people annotated ($N_v=1$ in our case), $Frag_i$ is the number of times a new ID is assigned to the target. Larger is better. 

$M3$, the Localization Error, evaluates the ability to precisely localize the target in the robot space. It consists of the average Euclidean distance between the target location estimated by the tracker and the ground truth location: 



\begin{equation}
M3(v) =  \frac{1}{N_v}\sum_{i=1}^{N_v} \sqrt{(X_i - X_{GT,i})^2 +  (Y_i - Y_{GT,i})^2} \label{eq:M3}
\end{equation}

where $(X_{GT,i}, Y_{GT,i})$ is the ground truth location of the target and $(X_{i}, Y_{i})$ is the predicted location on frame $i$. $N_v$ is the number of annotated frames for video $v$. Smaller values are better. 

\subsection{Results and Discussions}

\subsubsection{Performance}
\label{performance}

Since our target is an application in guiding robotics, our goal is a system with good accuracy in tracking a target subject, while being able to run in real-time on low-end GPUs like the ones can find in a laptop.
To assess the potentiality of our method, we also evaluated the three metrics against \textit{MPLT}, a state-of-the-art approach in panoramic video tracking introduced in Sec. \ref{people_tracking}. The results are shown in Table \ref{tab:Metrics}. 


\setlength{\tabcolsep}{6pt}
\begin{table}[h]
     \caption{The following tables collect the results from each tracking metric. We show the performance in each video and the average over the whole dataset. The processing speed on a GTX 1650 is shown too. Best results are in bold.}
     \label{tab:Metrics}
    \begin{subtable}[h]{0.45\textwidth}
        \centering
        \caption{M1, Normalized Tracking Time}
       \label{tab:M1}
        \begin{tabular}{c | c c c}
       \textbf{Video ID}  &  \textbf{TILES} & \textbf{ROI} & \textbf{MPLT}\\
        \hline \hline
vid1 & 0.97 & 0.97 & \textbf{0.98}\\ 
vid2 & 0.84 & \textbf{0.99} & 0.32\\ 
vid3 & \textbf{0.99} & \textbf{0.99} & 0.63\\ 
vid4 & \textbf{0.98} &\textbf{ 0.98} & 0.53\\ 
vid5 & \textbf{0.99} & \textbf{0.99} & 0.91\\ 
vid6 & \textbf{1.00} & 0.96 & 0.90\\ 
vid7 & 0.93 & 0.94 &\textbf{ 0.99}\\ 
vid8 & \textbf{0.96} & 0.92 & 0.90\\ 
vid9 & \textbf{0.97} & 0.96 & 0.95\\ 
vid10 & \textbf{0.98} & 0.96 & 0.96\\ 
        \hline
        \textbf{Average} & 0.96 &\textbf{ 0.97} & 0.81\\
       \end{tabular}
    \end{subtable}
    \hfill
    \begin{subtable}[h]{0.45\textwidth}
        \centering
        \caption{M2, Tracking Turnover}
       \label{tab:M2}
        \begin{tabular}{c | c c c}
        \textbf{Video ID}  &  \textbf{TILES} & \textbf{ROI} & \textbf{MPLT}\\
        \hline \hline
vid1 &\textbf{ 1.00} & \textbf{1.00} & \textbf{1.00}\\ 
vid2 & 0.50 & \textbf{1.00} & 0.50\\ 
vid3 & \textbf{1.00} & \textbf{1.00} &\textbf{ 1.00}\\ 
vid4 & 0.50 & \textbf{1.00} & 0.50\\ 
vid5 & \textbf{1.00} & \textbf{1.00} & \textbf{1.00}\\ 
vid6 &\textbf{ 1.00} & \textbf{1.00} & \textbf{1.00}\\ 
vid7 & \textbf{0.50} & 0.33 & \textbf{0.50}\\ 
vid8 & \textbf{0.50} &\textbf{ 0.50} & \textbf{0.50}\\ 
vid9 & \textbf{0.50} & \textbf{0.50} &\textbf{ 0.50}\\ 
vid10 & \textbf{1.00} & \textbf{1.00} & \textbf{1.00}\\ 
        \hline
        \textbf{Average} & 0.75 &\textbf{ 0.83} & 0.75\\
       \end{tabular}
    \end{subtable}
    
    \centering
    

\begin{subtable}[h]{0.65\textwidth}
        \centering
        \caption{M3, Localization Error}
       \label{tab:M3}
        \begin{tabular}{ c | c  c  c  }
       \textbf{Video ID} &\textbf{TILES} & \textbf{ROI} &\textbf{MPLT}\\ 
        \hline \hline
vid1 & \textbf{0.077} & 0.087  & 0.178 \\ 
vid2 & \textbf{0.394} & 0.408 & 0.496 \\ 
vid3 & 0.341 & \textbf{0.326} & 0.344 \\ 
vid4 & 0.465 &\textbf{ 0.406}  & 0.312 \\ 
vid5 &\textbf{ 0.121} & 0.13 & 0.373 \\ 
vid6 & 0.305 &\textbf{ 0.248} & 0.402 \\ 
vid7 & 0.261 & \textbf{0.224} & 0.149 \\ 
vid8 & 0.271 & \textbf{0.238} & 0.653\\ 
vid9 & \textbf{0.241} & 0.244 & 0.362 \\ 
vid10 & 0.296 & \textbf{0.254} & 0.652\\ 
        \hline
\textbf{Average} & 0.277 & \textbf{0.256} & 0.392 \\ 
        \end{tabular}
    \end{subtable}
    \begin{subtable}[h]{0.25\textwidth}
        \centering
       \caption{Processing speed}
        \label{tab:fps}
        \begin{tabular}{c | c}
        \textbf{System}  &  \textbf{FPS}\\
        \hline \hline
        TILES & 13.7\\ 
        ROI & \textbf{18.7}\\
        MPLT & 1.12\\
        \end{tabular}
    \end{subtable}
\end{table}

\begin{figure}[h]
\centering
    \includegraphics[width=0.92\textwidth]{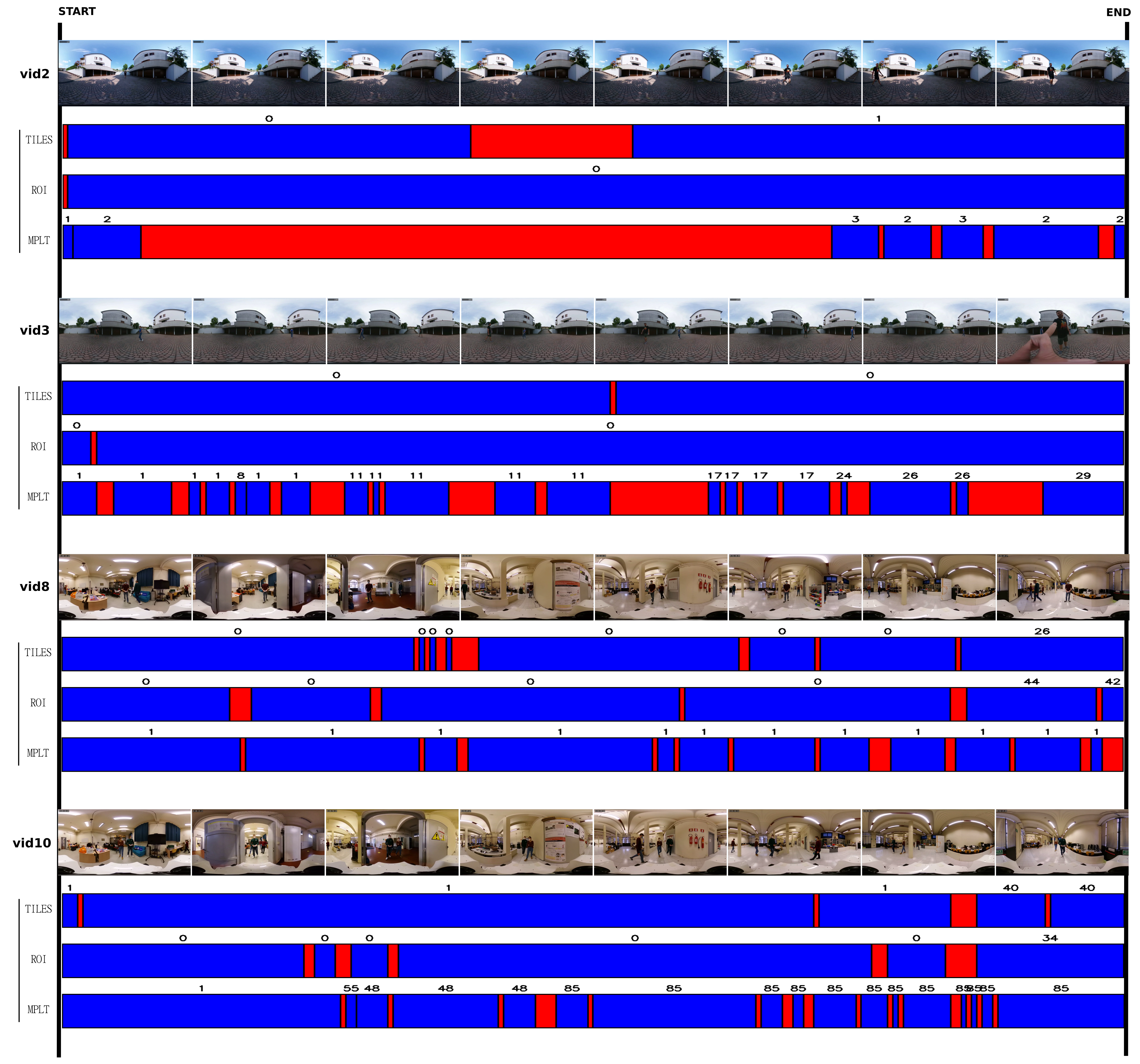}
    \caption{A visual representation of the performance of the systems in 4 samples from our dataset. Blue means the target is correctly tracked, red instead the target is lost. Above each bar, we indicate the current ID assigned by the tracker.}
    \label{fig:bars}
\end{figure}

From the Normalized Detection Time (M1) results in Table \ref{tab:M1}, it is possible to evaluate the strength of the detection capabilities. 
It is worth noting that \textit{MPLT} performs poorly in the sequences $vid2,3$ and $ 4 $. These sequences were recorded outdoor, letting the target go very far from the camera, up to 7.5 meters. The higher distance, the smaller the person size with respect to the rest of the image. \textit{ROI} and \textit{TILES} detectors, on the other hand, remain reliable even when the target is very far from the camera, with slightly better results for ROI. The advantage is probably due to the fact in ROI, the detector can focus in a smaller region which means less noise around the person. The Tracking Turnover (M2) in Table \ref{tab:M2} instead shows the capability of the system to keep track of the target. The difference between the three approaches, in this case, is less noticeable. \textit{MPLT} uses a traditional Kalman Filter and visual features to track, while our proposals use the UKF which is often more robust for modeling non-linear functions \cite{Vandyke2004}. \textit{MPLT} is also penalized by the poorer results in M1. For a better understanding of the performance in tracking, we also provide a visual representation in Fig. \ref{fig:bars} that confirms the previous considerations. In outdoor, where people can move further from the camera, \textit{MPLT} is more inaccurate and unstable. In indoor, the differences are less marked. The Localization Error (M3) in Table \ref{tab:M3} shows the accuracy of the systems in estimating the position of the target in the robot space. The results are clearly in favor of our systems. While \textit{MPLT} just uses the center of the detection as a reference point to estimate the distances assuming the height of a person is almost constant, our approach is using three joints and continuously updating the person height estimation. \textit{ROI} and \textit{TILES} contribute in providing an higher accuracy in human pose detection which is critical for an accurate localization in the robot space. Especially \textit{ROI} provides an average accuracy of around 0.25 meters, which is 65\% more precise than \textit{MPLT}. 

The Table \ref{tab:fps} shows the results for the processing speed. Despite we ran the tests on a low-end GPU (an NVIDIA GTX 1650), the results show that our methods overcome the compared method in processing speed, running between 10 and 20 times faster and reaching frame rates suitable for a robotic system with real-time constraints. The parallel inferences for the human pose detection - which is the most demanding task in the pipeline in terms of processing time - enables to speed up the elaboration of high-resolution images. \textit{ROI} is 5 fps faster than \textit{TILES} since it performs only two inferences in parallel, on smaller images. On the other hand, \textit{TILES} treats equally every person in the scene and it is more suitable for multi-person tracking. 
It is worth noticing that \textit{MPLT} is using a ResNet-50 \cite{he2015deep} as backbone, while our approaches are based on a lighter MobileNetV2 \cite{sandler2019mobilenetv2}. Also the tracking step in \textit{MPLT} takes more time, around 12 ms per frame, while our method takes around 1 ms.

\subsubsection{The impact of fisheye distortion}

\begin{figure}[h]
    \centering
    \includegraphics[width=0.7\textwidth]{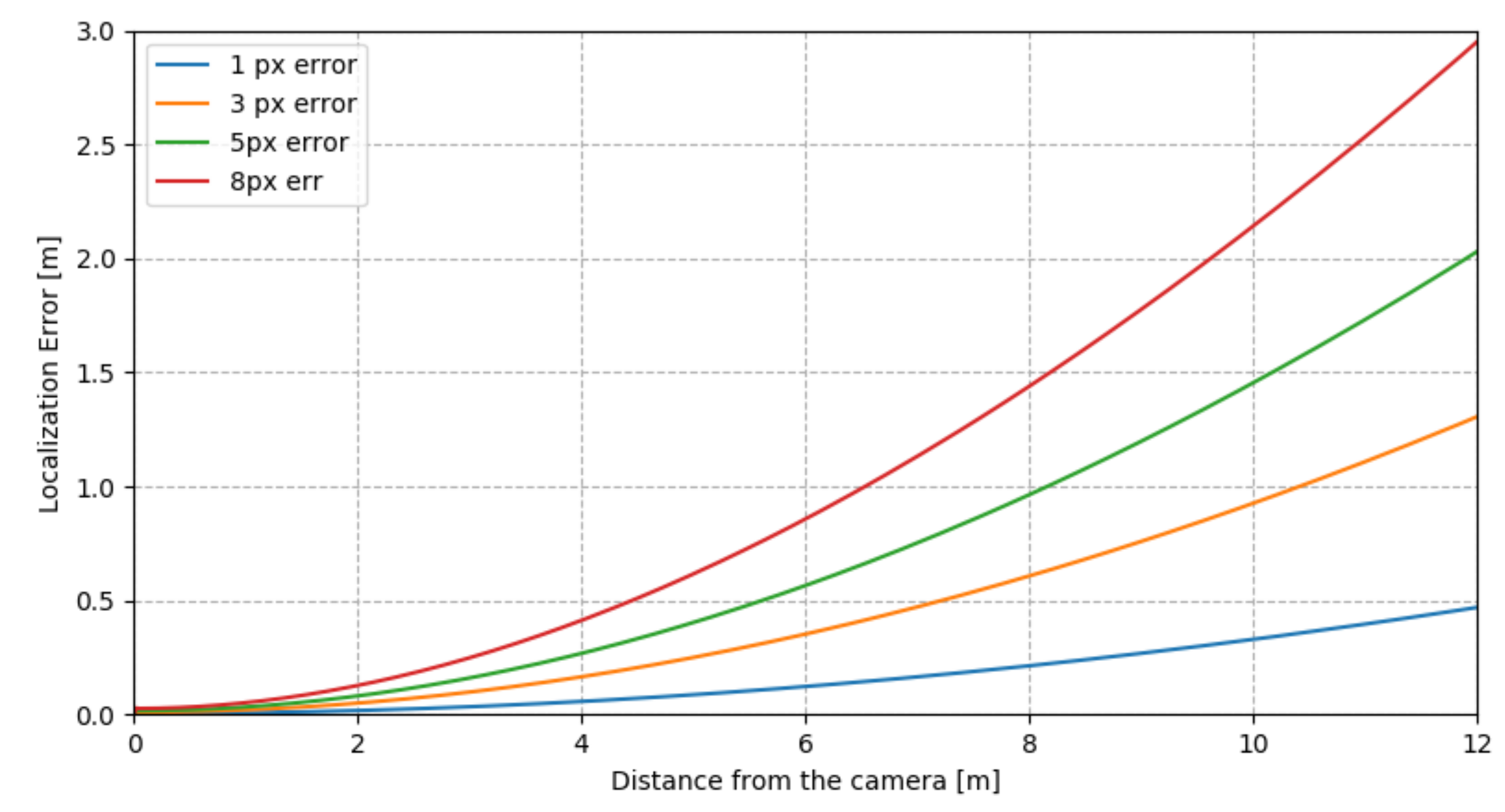}
    \caption{Localization Error vs Distance from the camera, with different localization error in the image space.}
    \label{fig:quant}
\end{figure}

The major problem in low-cost omnidirectional cameras is related to the fisheye lens. This type of lens allows to obtain a wider field of view while reducing the number of sensors in the camera but, at the same time, it introduces strong distortions. It is worth studying the impact of these distortions. 
The main problem is a rapid decrease in the size of objects with distance from the camera. In addition to making detection more hard, the accuracy of the localization is also reduced. As reported in Fig. \ref{fig:quant}, small errors in joint detection lead to large localization errors as the distance from the camera increases.

\begin{figure}[h!]
    \centering
    \includegraphics[width=0.85\textwidth]{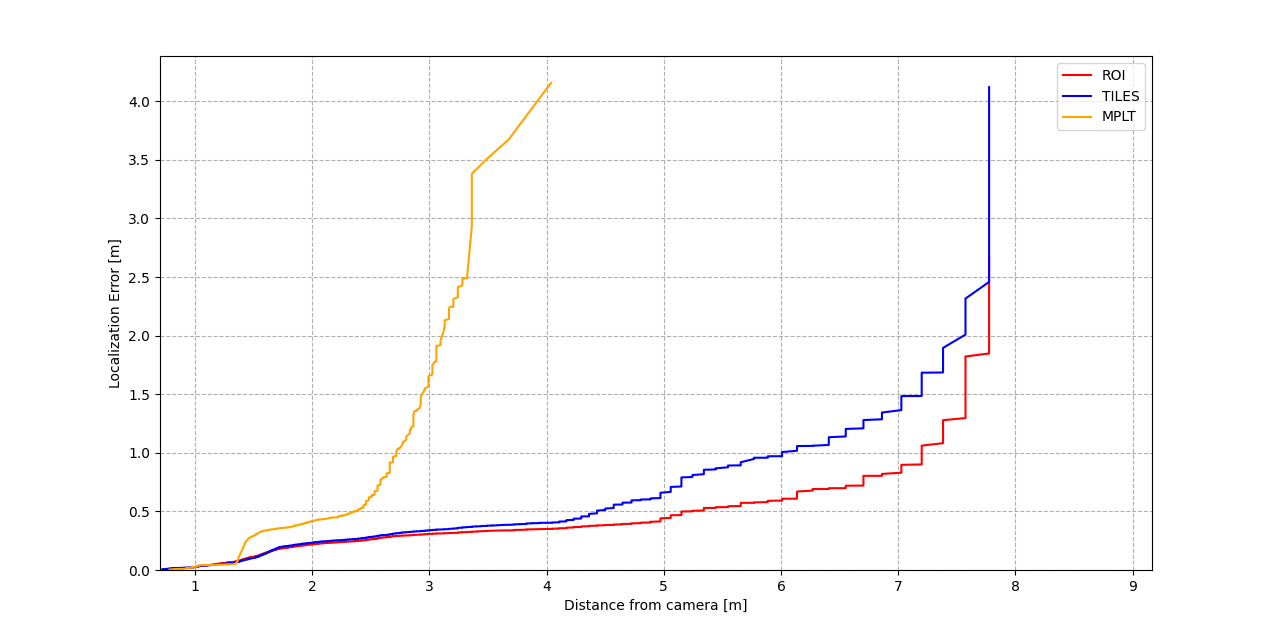}
    \caption{Localization Error vs Distance from the camera in different systems.}
    \label{fig:m5_full}
\end{figure}

The higher detection accuracy provided by our methods has a twofold benefit, as shown in Fig. \ref{fig:m5_full}. While \textit{MPLT} fails to detect subjects further than 4 meters, we can detect people 7-8 m away from the camera. More interesting, the more accurate joint estimation in the image provided by \textit{ROI} and \textit{TILES} can mitigate the growth of the localization error. With \textit{MPLT}, the error rapidly increases and overcomes 1 m when the person is 2.5 m away from the camera. Our methods are able to maintain the error under 1 m until 6 m for \textit{TILES} and 7 m for \textit{ROI}. This difference is also due to the camera model used in \textit{MPLT}, which probably cannot handle the strong distortion of fisheye lenses. The advantage of \textit{ROI} on \textit{TILES} is explainable with the focus in the target region, as discussed above. 

\section{Conclusions and Future Work}
\label{conclusions}

We have proposed an extension of a previous work on people detection and tracking, suitable to work with low-cost commercial omnidirectional cameras and targeting the application of guiding robotics. We introduced two algorithms, \textit{TILES} and \textit{ROI}, to tackle the challenges in panoramic videos such as high resolutions and fisheye distortions. Both the proposed algorithms overcome the current state-of-the-art in terms of detection accuracy, localization accuracy, and real-time processing according to our evaluations. \textit{ROI} shows higher performances both in accuracy and speed, but it is optimized for a single target person. In fact, it is specifically designed to work on guiding robots, which usually take one person as a reference. \textit{TILES} instead, with a very limited deterioration in performance, is a more generic approach, able to track multiple people with the same accuracy as the target. Moreover, the modified UKF and GNN allow us to solve the problems related with the \textit{wrap-around} to achieve an higher tracking stability. Since there is no large availability of panoramic video datasets, we also collected and labeled a dataset to evaluate our proposal against a state of art framework for people detection ad tracking in panoramic videos.

This work has demonstrated that is possible to easily adapt a traditional framework for people detection and tracking to work with panoramic videos. The use of low-cost omnidirectional cameras can be very effective on robots, where the small FoV often limits the system's capabilities. A 360$\degree$ FoV is particularly suitable when dealing with dynamic obstacles and targets like people, who can easily exit the FoV of traditional cameras. Given that, as future work we would like to do more extensive tests with a larger dataset, containing multi-person annotations. Relaxing the assumption that the camera is placed at a fixed height and orthogonal to the ground is desirable to enable larger and quicker adoption of our system. Further studies will involve application on a real guiding robotic system, aiming to use both the information of the target and the other people to manage the navigation of the robot. 

\section{Acknowledgements}

This research  was  partially  supported by MIUR  (Italian  Minister  for Education)  under  the initiative  ``Departments  of  Excellence"  (Law  232/2016)

\bibliographystyle{splncs}
\bibliography{biblio}

\end{document}